\documentclass[journal]{IEEEtran}

\usepackage{booktabs}
\usepackage{caption}
\usepackage{subcaption}
\usepackage{cite}
\usepackage{hhline}
\usepackage[cmex10]{amsmath}
\usepackage{graphicx}
    \graphicspath{{./figs/}}
    \DeclareGraphicsExtensions{.pdf}
\usepackage{balance}
\usepackage{algorithmic}
\usepackage{dsfont}
\usepackage{hhline}
\usepackage[usenames]{color}
\usepackage{multirow}
\usepackage{amssymb}
\usepackage{array}
\usepackage{url}
\usepackage{xspace}
\usepackage{mathtools}

\usepackage[table,xcdraw]{xcolor}
\usepackage{blindtext}
\usepackage{diagbox}
\usepackage{cleveref}
\usepackage{tablefootnote}

\begin{document}
%
\title{Time Series Clustering With Random Convolutional Kernels}
%
%
%

\author{\IEEEauthorblockN{
Jorge Marco-Blanco\IEEEauthorrefmark{1},
Rubén Cuevas\IEEEauthorrefmark{1}\IEEEauthorrefmark{2}
}

\IEEEauthorblockA{\IEEEauthorrefmark{1} Universidad Carlos III de Madrid, Spain}\\
\IEEEauthorblockA{\IEEEauthorrefmark{2}UC3M-Santander Big Data Institute, Spain}\\


}
\maketitle

\begin{abstract}
Time series data, spanning applications ranging from climatology to finance to healthcare, presents significant challenges in data mining due to its size and complexity. One open issue lies in time series clustering, which is crucial for processing large volumes of unlabeled time series data and unlocking valuable insights. Traditional and modern analysis methods, however, often struggle with these complexities. To address these limitations, we introduce R-Clustering, a novel method that utilizes convolutional architectures with randomly selected parameters. Through extensive evaluations, R-Clustering demonstrates superior performance over existing methods in terms of clustering accuracy, computational efficiency and scalability. Empirical results obtained using the UCR archive demonstrate the effectiveness of our approach across diverse time series datasets. The findings highlight the significance of R-Clustering in various domains and applications, contributing to the advancement of time series data mining.
\end{abstract}

\IEEEpeerreviewmaketitle

\section{Introduction}
\label{sec:introduction}

Time is a natural concept that allows us to arrange events in a sorted way from the past to the future in an evenly distributed manner (such as days, seasons of the year or years). Data points are usually taken at successive equally spaced points in time. Based on this way of describing events, the concept of time series as a sequence of discrete data points evenly distributed arises naturally. From this perspective, time series can describe a wide range of natural and social phenomena that evolve in time. A few samples are climate and weather trends, seismic measures, stock prices, sales data, biomedical measurements, body movements or Internet traffic \cite{aghabozorgi2015time}. The study of time series involves identifying its statistical properties, including trends, seasonality, and autocorrelation, and using this information to build models for prediction and classification.

According to \cite{yang200610}, who reviewed the work of the most active researchers in data mining and machine learning for their opinions, the problem of sequential and time series data is one of the ten most challenging in data mining. Its most relevant applications are prediction and classification (supervised or unsupervised). As for time series data availability, the significant advances in data storage, collection and processing produced during the last decades have made available a large amount of data which is growing exponentially. Unfortunately, most of this data remains unlabelled, making it useless for some critical applications such as supervised learning.

With the ever-increasing growth rate of available data, we need scalable techniques to process it. Clustering serves as a solution for extracting valuable insights from unlabelled data, allowing a vast quantity of this data to be processed efficiently. Clustering can be defined as the task of grouping a set of objects into different subsets, or clusters, where the objects in each subset share certain common characteristics. Since it does not require human supervision or hand-labeling, clustering is classified within the domain of unsupervised learning techniques. This makes clustering particularly suited to keeping pace with the influx of available data.

Although time series are usually considered a collection of data points sorted chronologically, they can also be regarded as a single object \cite{kumar2006time} and, therefore, subject to clustering. In scenarios involving significant volumes of time series data, clustering is especially helpful in discovering patterns. In the case of rare patterns, it facilitates anomaly and novelty detection, while for frequent patterns, it aids in prediction and recommendation tasks \cite{aghabozorgi2015time}. Examples of such applications include detecting web traffic anomalies in Informatics \cite{lakhina2005mining} and gene classification in Biology \cite{mcdowell2018clustering}. Time series clustering can also serve as a pre-processing technique for other algorithms such as rule discovery, indexing, or classification \cite{aghabozorgi2015time}.


Despite the crucial role of time series clustering across diverse fields, existing approaches often struggle with the complexity of high-dimensional, noisy real-world time series data. Furthermore, while automated feature extraction methods have shown success, they require more parameters, data, and longer training periods, as discussed in Section \ref{sec:relatedWork}. There is, therefore, a need for more efficient, accurate, and scalable methods for time series clustering.

In this paper, we make several contributions to the field of time series clustering. In Section \ref{sec:methods}, we introduce R-Clustering, a novel time series clustering algorithm that uses convolutional architectures with static, randomly selected kernel parameters. This approach addresses the challenge of scalability and resource-intensive training, prevalent in current methods. Subsequently, in Section \ref{sec:results} we provide a comprehensive evaluation of R-Clustering, benchmarking its performance against state-of-the-art methods through the use of the UCR archive and detailed statistical analyses.  We contrast R-Clustering against eight other reference clustering algorithms across 72 diverse time series datasets. Remarkably, R-Clustering outperforms the other algorithms in 33 out of the 72 datasets, with the second-best performing algorithm leading in only 13 datasets. In addition, R-Clustering achieves the highest mean clustering accuracy and rank, and is also the fastest algorithm across the datasets evaluated. We further demonstrate its scalability for larger datasets. These findings highlight the superior accuracy and scalability of R-Clustering, emphasizing its potential for deployment in large-scale applications.

\section{Related Work}
\label{sec:relatedWork}
In this section, we will first review the most relevant methods for time series clustering. Then, our focus will shift to feature extraction methods applied to 2D and 1D data: this discussion will cover general applications of these methods and make a distinction between those methods that use learnable kernels and those that use static kernels. To conclude, we will explore the potential application of static kernel methods to time series clustering

\subsection{Different Approaches to Time Series Clustering}

Different approaches to the clustering of time series, according to \cite{aghabozorgi2015time}, can be classified in three groups: model-based, shape-based and feature-based. In the model-based methods, the time series are adjusted to a parametric model; since the parameters do not have a time series structure, a universal clustering algorithm is applied. Some popular parametric models are autorregresive-moving-average (ARMA) models or the Hidden Markov Model. However, these techniques might not accurately model high-dimensional and noisy real-world time-series \cite{langkvist2014review}. Existing methods that may address the noise issue, such as wavelet analysis or filtering, often introduce lag in the processed data, reducing the prediction accuracy or the ability to learn significant features from the data \cite{yang200610}.

In the shape-based approach, the clustering algorithm is applied directly to the time series data with no previous transformation. Unlike to the model-based approach, these methods employ a clustering method equipped with a similarity measure appropriate for time series. For instance, \cite{paparrizos2015k} introduce the k-shape algorithm with a shape-based approach. In this work, the authors suggest that feature-based strategies may be less optimal due to their domain dependence, requiring the modification of the algorithms for different datasets. A different approach in shape-based clustering involves utilizing a universal clustering algorithm, such as the widely used k-means algorithm \cite{macqueen1967classification}, along with a suitable distance measure that considers the unique characteristics of time series data, such as dynamic time warping (DTW), introduced by \cite{berndt1994using}. While DTW has been shown to perform better for time series clustering, it comes with the downside of having a time complexity of \begin{math} \mathcal{O}(n^2) \end{math}, where n is the length of time series. Compared to DTW, the Euclidean distance, which is the distance usually used in the k-means algorithm \cite{likas2003global}, has a complexity of \begin{math} \mathcal{O}(n) \end{math} \cite{jain1999data}.

Feature-based methods first extract relevant time series features and later apply a conventional clustering algorithm \cite{aghabozorgi2015time}. As we will see in the next section, feature-based algorithms have been proven quite successful for image clustering and classification. According to \cite[p.~1798]{bengio2013representation}, the performance of machine learning algorithms relies strongly on the correct choice of feature representation.


\subsection{Feature extraction}
The incorporation of expert knowledge has been shown to improve machine learning performance, as demonstrated by previous research such as \cite{bengio2013representation} and \cite{sebastiani2002machine}. Traditionally, this has been achieved through the manual design of feature extractors based on the specific task. Examples of such strategy include Lowe's algorithm  \cite{lowe1999object} for image feature extraction using hand-designed static filters to induce specific invariances in the transformed data. Also, it is well known that the discrete Gaussian kernel can be used to perform a wide range of filtering operations, such as low-pass filtering, high-pass filtering, and band-pass filtering \cite{kailath1980linear}. In the context of digital signal processing, FIR (Finite Impulse Response) filters operate by applying a weight set, or "coefficients," to a selection of input data to calculate an output point, essentially a discrete convolution operation \cite{proakis1996digital}. It is important to remark, however, that this process requires manual calibration of the filter to selectively allow certain frequencies through while suppressing others.These techniques, while effective, present scalability limitations due to their dependency on human supervision for the design process. As data volumes continue to grow exponentially, there is an increasing need for automated feature extraction techniques that can efficiently handle large datasets without the extensive requirement for expert knowledge.


In some domains, such as the 2D shape of pixels in images or time series having a 1D structure, it is possible to learn better features using only a basic understanding of the input as an alternative to incorporating expert knowledge. In images and time series data, patterns are likely to reproduce at different positions, and neighbours are likely to have strong dependencies. The existence of such local correlations is the basis for applying a local feature extractor all over the input and transforming it into a feature map of a similar structure. A convolution operation usually performs this task with a convolutional kernel \cite[p.~1820]{bengio2013representation}. The use of local feature extractors has several benefits in machine learning. First, it allows for the efficient extraction of relevant features from large datasets without requiring expert knowledge. Second, it can improve the accuracy of machine learning models by capturing local correlations in the input data. Finally, it can reduce the dimensionality of the input data, making it easier to process and analyze.

The academic community has achieved substantial advances during the last decade applying this technique to the problems of image classification \cite{krizhevsky2012imagenet}, time series forecasting \cite{greff2016lstm} or time series classification \cite{ismail2019deep} among others. In the field of time series clustering, \cite{ma2019learning} have developed a convolutional model with deep autoencoders. 


\subsubsection{Feature extraction with learnable kernels}

In most convolution algorithms, kernel weights are typically learnt during the training process, as described in \cite{bengio2013representation}. Over the last few years, the academic community has been actively exploring the applications of such convolutional in tasks such as image classification and segmentation. This is illustrated by works like \cite{ciresan2011flexible}, \cite{pereira2016brain}, and \cite{krizhevsky2012imagenet}. These models use convolutional layers in neural networks to transform image data, refining and compressing the original input into a compact, high-level feature representation. The architecture then applies a linear classification algorithm on this condensed feature representation as the final step.

Some authors have applied convolutional architectures to the problem of image clustering, as described in \cite{caron2018deep} and  \cite{xie2016unsupervised}. In both models, a network transforms input data into an enhanced feature representation, followed by a clustering algorithm. This dual optimization approach adjusts both the network weights and the clustering parameters simultaneously. The first model predicts input labels based on the clustering results, then the error in these predictions is backpropagated through the network  to adjust the network parameters. The second model utilizes an autoencoder with a middle bottleneck. The network parameters are first optimized using the reconstruction loss. Then, the decoder section of the architecture is discarded and the output features from the bottleneck form the new data representation. This condensed data representation is then processed by a clustering algorithm.

Convolutional neural networks (CNNs) have also become increasingly popular in the field of time series analysis due to their ability to capture local patterns and dependencies within the data. For example, \cite{wang2017time} proposed a model that uses a combination of 1D and 2D CNNs to extract both temporal and spatial features from multivariate time series data. Another model, proposed by \cite{zhao2017convolutional}, uses a dilated convolutional neural network to capture long-term dependencies in time series data. More recently, \cite{ismail2019deep} used deep convolutional neural networks for time series classification achieving state-of-the-art results. Despite the success of CNN-based models in time series analysis, there are still challenges that need to be addressed. One challenge is the selection of appropriate hyperparameters, such as the number of filters and the filter size, which can greatly affect the performance of the model.

In recent work, \cite{ma2019learning} proposed a novel clustering algorithm for time series data that builds upon previous image clustering techniques. The model leverages an autoencoder to obtain a reconstruction loss, which measures the difference between the original time series and its reconstructed version. Simultaneously, the model employs a clustering layer to predict the labels of the time series, which is used to compute a prediction loss. By combining both losses, the model jointly optimizes the parameters of the network and the clustering parameters. This approach allows for more accurate clustering of time series data, as it takes into account both the reconstruction error and the predicted labels.


Deep learning faces considerable challenges due to the complexity of its models, which typically involve multiple layers and a significant number of parameters \cite{goodfellow2016deep}, \cite{bengio2013representation}. This complexity results in practical issues, including the requirement for substantial computational resources and vast amounts of training data. There is also the potential for overfitting and the necessity for fine tuning to optimize the model's performance.

When deep learning is applied to time series analysis, further complications arise. Time series data exhibit unique characteristics like temporal dependencies and seasonal patterns, requiring specialized treatment as indicated by \cite{wang2017time} and \cite{ismail2019deep}. In addition to the challenges associated with deep learning models, time series data also pose unique challenges due to their often high-dimensional and highly variable nature \cite{langkvist2014review}. This can make it difficult to select appropriate hyperparameters and optimize the model's performance.

\subsubsection{Feature extraction with static random kernels}

Convolutional models with static parameters offer a distinct approach to feature extraction. Instead of learning the weights of the convolutional filters during the training process, these models use fixed or static parameters, resulting in faster computation times and simpler architectures \cite{bengio2013representation}. The work of \cite{huang2014insight} and \cite{jarrett2009best} supports this approach, demonstrating that convolutional models with weights that are randomly selected, or random kernels, can successfully extract relevant features from image data. Additionally, \cite{saxe2011random} argue that choosing the right network design can sometimes be more important than whether the weights of the network are learned or random. They showed that convolutional models with random kernels are frequency selective and translation invariant, which are highly desirable properties when dealing with time series data.

\cite{dempster2020rocket} provide evidence that convolutional architectures with random kernels are effective for time series analysis. The authors of the paper demonstrate that their proposed method, called ROCKET (Random Convolutional KErnel Transform), achieves state-of-the-art accuracy on several time series classification tasks while requiring significantly less computation than existing methods. The authors further improved the efficiency of their method by introducing MiniRocket \cite{dempster2021minirocket}, an algorithm that runs up to 75 times faster than ROCKET on larger datasets, while maintaining comparable accuracy


Motivated by the success of random kernels applied to the problem of feature extraction of time series and image data, we propose a simple and fast architecture for time series clustering using random static kernels. To the best of our knowledge, this is the first attempt to use convolutional architectures with random weights for time series clustering. This approach eliminates the need for an input-reconstructing decoder or a classifier for parameter adjustments, commonly found in previous time series clustering works.Instead, the method applies the convolution operation with random filters over the time series input data to obtain an enhanced feature representation, which is then fed into a K-means clustering algorithm. By eliminating the need for a reconstruction loss or classification loss, our method is more efficient and easier to implement than existing methods for time series clustering. We have conducted several experiments to test the effectiveness and scalability of the algorithm, and our results show that it outperforms current state-of-the-art methods. As such, our research provides a substantial contribution to the field of time series clustering, offering a promising new avenue for further advancements.

\section{Methodology}
\label{sec:methods}

This section explains the proposed clustering algorithm and the evaluation methods, including a statistical analysis. Regarding data, we use the UCR archive \cite{dau2019ucr}, that is a vital tool for time series researchers, with over one thousand papers employing it. At the time of writing this work, it consists of 128 datasets of different types: devices, ECGs, motion tracking, sensors, simulated data and spectrographs.

\subsection{Clustering algorithm}

We introduce \emph{R-clustering}, a new algorithm for time series clustering, which is composed of three stages of data processing elements connected in series. The first element is a feature extractor, comprised of a single layer of static convolutional kernels with random values. This extractor transforms the input time series into an enhanced data representation. The second processing element employs principal component analysis for dimensionality reduction, selecting combination of features that account for the majority of the variance in the input data. The last element is a K-means algorithm \cite{macqueen1967classification} that utilizes Euclidean distance, thereby providing the algorithm with its clustering capabilities. Next we describe each of this elements in detail.

\subsubsection{Feature extraction}
\label{subsec:feature_extraction}

For the feature extractor, we propose a modified version of the one used in Minirocket \cite[p.~251]{dempster2021minirocket} which is based on a previous algorithm called Rocket \cite{dempster2020rocket}. In particular, we employ randomly selected values for the bias and adjust the configuration of the hyperparameters to better suit the distinct challenge of clustering. We have conducted an optimization process of the hyperparameters (kernel length and number of kernels in this case). To avoid overfitting the UCR archive we have chosen a development set of 36 datasets, the same group of datasets used in the original algorithm (Rocket)

Our feature extractor is composed of 500 kernels, each of length 9, optimized for clustering accuracy as the result of a hyperparameter search detailed in Section: \ref{hyperparameter}. The kernel weights are restricted to either 2 or -1. As \cite{dempster2021minirocket} demonstrated, constraining the weight values does not significantly compromise accuracy, but it substantially enhances efficiency.

Dilations represent another significant parameter in the model. Dilation in 1D kernels refers to the expansion of the receptive field of a convolutional kernel in one dimension. This expansion is achieved by inserting gaps between the kernel elements, which increases the distance between the elements and allows the kernel to capture information from a larger area. This technique expands the receptive field of the kernel, allowing them to identify patterns at various scales and frequencies. Our model follows the configuration proposed by \cite{dempster2021minirocket}, which employs varying dilations adapted to the specific time series being processed, to ensure recognition of most potential periodicities and scales in the data. The number of dilations is a fixed function of the input length and padding.

Other relevant configuration parameter is the selection of the bias values, which are chosen as follows: first each kernel is convolved with a time series selected at random from the dataset. Then, the algorithm draws the bias values randomly from the quantiles of the result of the convolution. This process ensures that the scale of the bias values aligns with that of the output.

After the convolution of the input series with these kernels under the mentioned configuration, the proportion of positive values (PPV) of the resulting series is calculated. The transformed data consists of 500 features (the same as the number of kernels) with values between 0 and 1.  

A thorough examination of the original feature extractor stage reveals there are artificial autocorrelations, not from the time series data, produced by the implementation of the algorithm. This behaviour could affect the performance of the clustering stage of R-clustering, or of a future algorithm using this feature extractor, because it would likely detect these unnatural patterns. Also, these patterns could mask legit features of time series, obtaining misleading results. We have identified the origin of this issue in the selection of the bias values. For a time series $ X $ convolved with a kernel $W$  PPV is computed as the proportion of positive values of $(W*X - \text{bias})$ where $*$  denotes convolution. To determine the bias values, our  original algorithm selects a training instance $ X $ and calculate its convolution with the specified dilation d and kernel: $W_d$. The quantiles of the resulting convolution output are used as the bias values. In the implementation of the original algorithm, we identified manner in which the bias values where sorted produced the artificial autocorrelations. To rectify this, we have randomly permuted them, effectively removing the artificial autocorrelations. (see \ref{subsec:RK-transformation} for a detailed description of the output of the feature extractor after and before the modification).


\subsubsection{Dimensionality reduction with Principal Component Analysis}

The "curse of dimensionality" can potentially impact the performance of the K-means clustering in the third stage of our algorithm, particularly given our high-dimensional context with 500 features, as detailed in the previous subsection. In high-dimensional space, the distance between data points becomes less meaningful, and the clustering algorithm may struggle to identify meaningful clusters \cite{beyer1999nearest}. This is because the distance between any two points tends to become more uniform, leading to the loss of meaningful distance metrics.  As the number of dimensions increases, the volume of the space increases exponentially, and the data becomes more sparse, making it difficult to identify meaningful clusters \cite{aggarwal2001surprising}. For these reasons, it is convenient to reduce the number of features to improve the performance of K-means clustering in a context of a high-dimensional space. In our particular case, due to the random nature of the kernel weights used in the convolutions with the input data, we expect that many components of the transformed data may not be significant. Hence, implementing a dimensionality reduction method can be beneficial in multiple ways. 

We propose using Principal Component Analysis (PCA), a dimensionality reduction technique that can identify the crucial dimensions or combinations of dimensions that account for the majority of data variability. Given that our problem is unsupervised, PCA is specially apt: it focuses on the inherent statistical patterns within the data, independently of any evaluation algorithm. As to why not using PCA directly to the time series data and skipping the convolution transformation, PCA does not take into account the sequential ordering of data, therefore it may not fully capture the underlying structure of time series.

One challenge associated with implementing PCA is determining the optimal number of principal components to retain, which will define the final number of features. Common techniques for this include the elbow method and the Automatic Choice of Dimensionality for PCA \cite{minka2000automatic}. The elbow method involves visualizing the explained variance as a function of the number of components, with the 'elbow' in the plot suggesting the optimal number. However, because this technique relies on visual interpretation, it may not be suitable for our automated algorithm. The second method, based on Bayesian PCA, employs a probabilistic model to estimate the optimal number of dimensions. This method, while powerful, might not always be applicable, particularly in the context of high-dimensional data like time series, where it may be challenging to satisfy the assumption of having more samples than features.

We therefore opt to determine the number components by analyzing the explained variance of the data introduced by each additional dimension until it is not very significant. We choose to consider increments of 1\% as no significant. As experiments from \cite{beyer1999nearest} show, the number of selected dimension lies between 10 and 20 which are the dimensions beyond which the effect of the curse of dimensionality can produce instability of an algorithm based on the Euclidean distance. In accordance with this result, our experiments show that most of the times the number of dimensions selected with our method are between 10 and 20.

In summary, we incorporate an additional stage into our algorithm that employs Principal Component Analysis (PCA) to reduce the dimensionality of the features prior to the implementation of the K-means algorithm.

\begin{figure*}[h]
    \centering
    \includegraphics[width=0.85\linewidth]{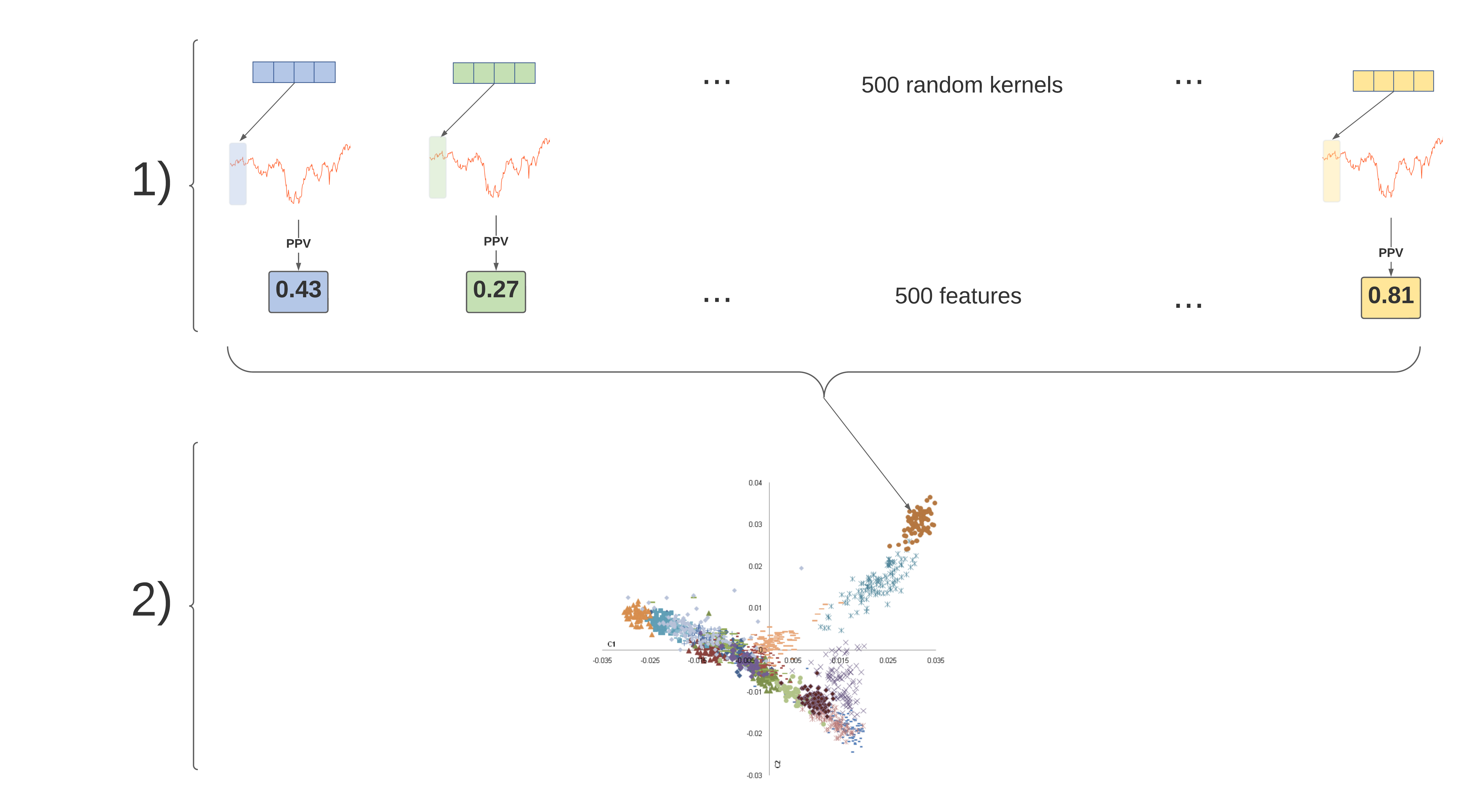}
    \caption{The figure illustrates the various steps involved in the R-clustering algorithm: 1)Initially, the input time series is convolved with 500 random kernels. Following this, the Positive Predictive Value (PPV) operation is applied to each the convolution results, generating 500 features with values spanning between 0 and 1. 2)The next phase involves applying Principal Component Analysis (PCA) for dimensionality reduction. This procedure results in a more manageable set of features, reducing the original 500 to between 10 and 20. Finally, the processed and dimensionality-reduced data are clustered using the K-means algorithm.}
    \label{fig:rkc-schema}
\end{figure*}

\subsubsection{K-means with Euclidean distance}

The findings in section \ref{subsec:RK-transformation}, demonstrating the absence of artificial autocorrelations in the output of the first stage, along with the dimensionality reduction via Principal Component Analysis (PCA) of the second stage, suggest that our algorithm's transformation considerably reduces the time series properties of the features following the first two stages.  This reduction simplifies the problem, making it more amenable to traditional raw data algorithms which are typically less complex and less demanding in terms of computational resources compared to algorithms designed specifically for time series data, such as those using Dynamic Time Warping (DTW) or shape-based distances. Consequently, in the third stage of R-clustering, we adopt a well-established clustering technique: the K-means algorithm \cite{macqueen1967classification} with Euclidean distance. This combination is widely recognized and has been extensively tested within the scientific community for clustering problems \cite{jain1999data}, \cite{likas2003global}. K-means partitions data into a number K (set in advance) of clusters by iteratively assigning each data point to the \emph{nearest} mean center (centroid) of a cluster. After each new assignment, the centroids are recalculated. When the training process finishes, the resulting centroids can be used to classify new observations. To evaluate the \emph{nearest} centroid, a distance metric must be defined. Using the Euclidean distance will result in a more efficient algorithm since it has a time complexity of \begin{math} \mathcal{O}(n) \end{math}. In contrast, using DTW as a distance metric would result in a time complexity \begin{math} \mathcal{O}(n^2) \end{math}. Figure \ref{fig:rkc-schema} provides a schema of the R-clustering algorithm and its stages.

\subsection{Evaluation method}
To the authors' knowledge, the only benchmark for time series clustering using the widely used UCR dataset is the one presented by \cite{javed2020benchmark}. This benchmark compares eight popular clustering methods that cover three categories of clustering algorithms (partitional, density-based, and hierarchical) and three distance measures (Euclidean, Dynamic time warping, and shape-based). Our evaluation of R-clustering's performance uses the same 112 datasets (36 development datasets and 76 validations datasets) from the UCR archive as the benchmark, with 16 out of the 128 total datasets omitted; 11 due to their variable lengths, and 5 because they comprise only one class. We ensure a fair and direct comparison by adhering to the same evaluation procedures defined in the benchmark study. The number of clusters for each dataset is known in advance since the UCR archive is labeled, and this number is used as an input for the clustering algorithms in the benchmark and the R-clustering algorithm. In case the number of clusters were not known, different methods exist to estimate them, such as the elbow method, but evaluating these methods is not part of the benchmark's paper or this paper.

Several metrics are available for the evaluation of a clustering process, such as Rand Index (RI) \cite{rand1971objective}, Adjusted Mutual Information \cite{vinh2009information} or Adjusted Rand Index (ARI) \cite{hubert1985comparing}. Among these, ARI is particularly advantageous because its output is independent of the number of clusters while not adjusted metrics consistently output higher values for higher number of clusters \cite{javed2020benchmark}. It is essentially an enhancement of the RI, adjusted to account for randomness. Additionally, \cite{steinley2004properties} explicitly recommends ARI as a superior metric for evaluating clustering performance. Based on these reasons and for comparability with the benchmark, we use the Adjusted Rand Index (ARI) to evaluate R-clustering. This choice also ensures compatibility with the benchmark study, facilitating meaningful comparisons of our results.

In comparing the performance of various algorithms, we adhere to the methods used in the benchmark study. Specifically, we calculate the following across the same 112 datasets: the number of instances where an algorithm achieves the highest Adjusted Rand Index (ARI) score, denoted as the 'number of wins'; the mean ARI score; and the mean rank of all algorithms. Results in subsection \ref{rclustering_results} indicate that R-clustering outperforms the other algorithms across all these measures, demonstrating its efficacy. We also conducted statistical tests to determine the significance of the results and considered any limitations or assumptions of the methods.




The problem of comparing multiple algorithms over multiple datasets in the context of machine learning has been treated by several authors \cite{demvsar2006statistical}, \cite{garcia2008extension}, \cite{benavoli2016should}. Following their recommendations, we first compare the ranks of the algorithms as suggested by \cite{demvsar2006statistical} and use the Friedman test to decide whether there are significant differences among them. If the test rejects the null hypothesis ("there are no differences"), we try to establish which algorithms are responsible for these differences. As \cite{garcia2008extension} indicates, upon a rejection of the null hypothesis by the Friedman test, we should proceed with another test to find out which algorithms produce the differences, using pairwise comparisons. Following the recommendations of the authors of the UCR Dataset,  \cite{dau2019ucr}, we choose the Wilcoxon sign-test for the pairwise comparisons between the R-clustering algorithm and the rest. As outlined in subsection \ref{rclustering_results}, we initially conduct comparisons between all benchmark algorithms and R-clustering, employing the latter as a control classifier to identify any significant differences. The results indicate that R-Clustering's superior performance compared to the other algorithms is statistically significant.

Additionally, to enhance the insights provided by the benchmark study and following the suggestions from \cite{garcia2008extension}, we carry out a new experiment that involves pairwise comparisons among all possible combinations within the set comprising of the benchmark algorithms and R-Clustering.


\subsection{Implementation and reproducibility}
We use Python 3.6 software package on Windows OS with 16GB RAM, Intel(R) Core(TM) i7-2600 CPU 3.40GHz processor for their implementation. In our study, we use a variety of reputable libraries that have been widely used and tested, therefore ensuring reliability. These include:

\begin{itemize}
\item \textit{sktime} \cite{loning2019sktime}, a standard Python library used for evaluating the computation time of the Agglomerative algorithm and for extracting data from the UCR dataset.
\item The code from \cite{paparrizos2015k}, which we utilize to evaluate the computation time of the K-shape algorithm.
\item \textit{scikit-learn} \cite{scikit-learn}, employed for executing the K-means stage of R-clustering and for evaluating the adjusted rand index.
\item Functions provided in \cite{ismail2019deep}, used for calculating statistical results.
\end{itemize}

We make our code publicly available \footnote{https://github.com/jorgemarcoes/R-Clustering} and base our results on a public dataset. This provides transparency and guaranteess full reproducibility and replicability of the paper following the best recommended practices in the academic community.

\section{Results}
\label{sec:results}

This section initiates with an investigation of the outputs generated during the feature extraction stage, as explicated in \ref{subsec:feature_extraction}. We continue with a search for optimal hyperparameters by examining various algorithm configurations. Following this, we present the results of applying the R-clustering algorithm to 112 datasets extracted from the UCR archive and perform a statistical analysis with other algorithm  evaluated in the benchmark study \cite{javed2020benchmark} using R-Clustering as a control classifier. Subsequently, we showcase the scalability results of R-Clustering. We conclude this section by conducting a comprehensive statistical comparison of all the benchmark algorithms, along with R-Clustering, thereby contrasting every single one against the rest.

\subsection{Analysis of the Feature Extraction Stage Output}
\label{subsec:RK-transformation}

\begin{figure}[h]
    \centering
    \includegraphics[width=\linewidth]{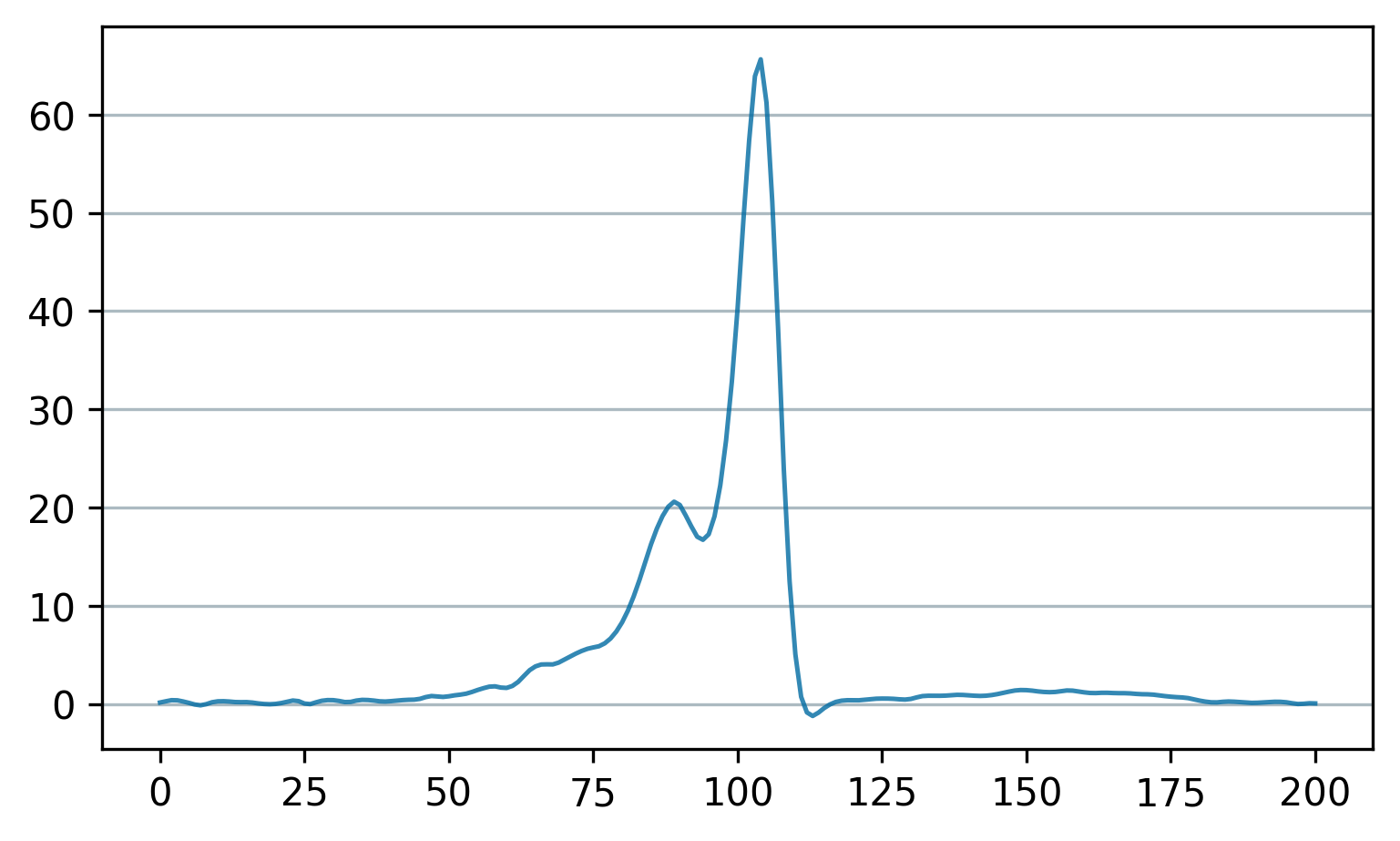}
    \caption{Sample time-series from Fungi dataset included in the UCR archive. The dataset contains high-resolution melt curves of the rDNA internal transcribed spacer (ITS) region of 51 strains of fungal species. The figure shows the negative first derivative (-dF/dt) of the normalized melt curve of the ITS region of one of such species}
    \label{fig:fungi-sample}
\end{figure}

\begin{figure*}[t]
  \begin{subfigure}[b]{0.47\linewidth}
    \includegraphics[width=\textwidth]{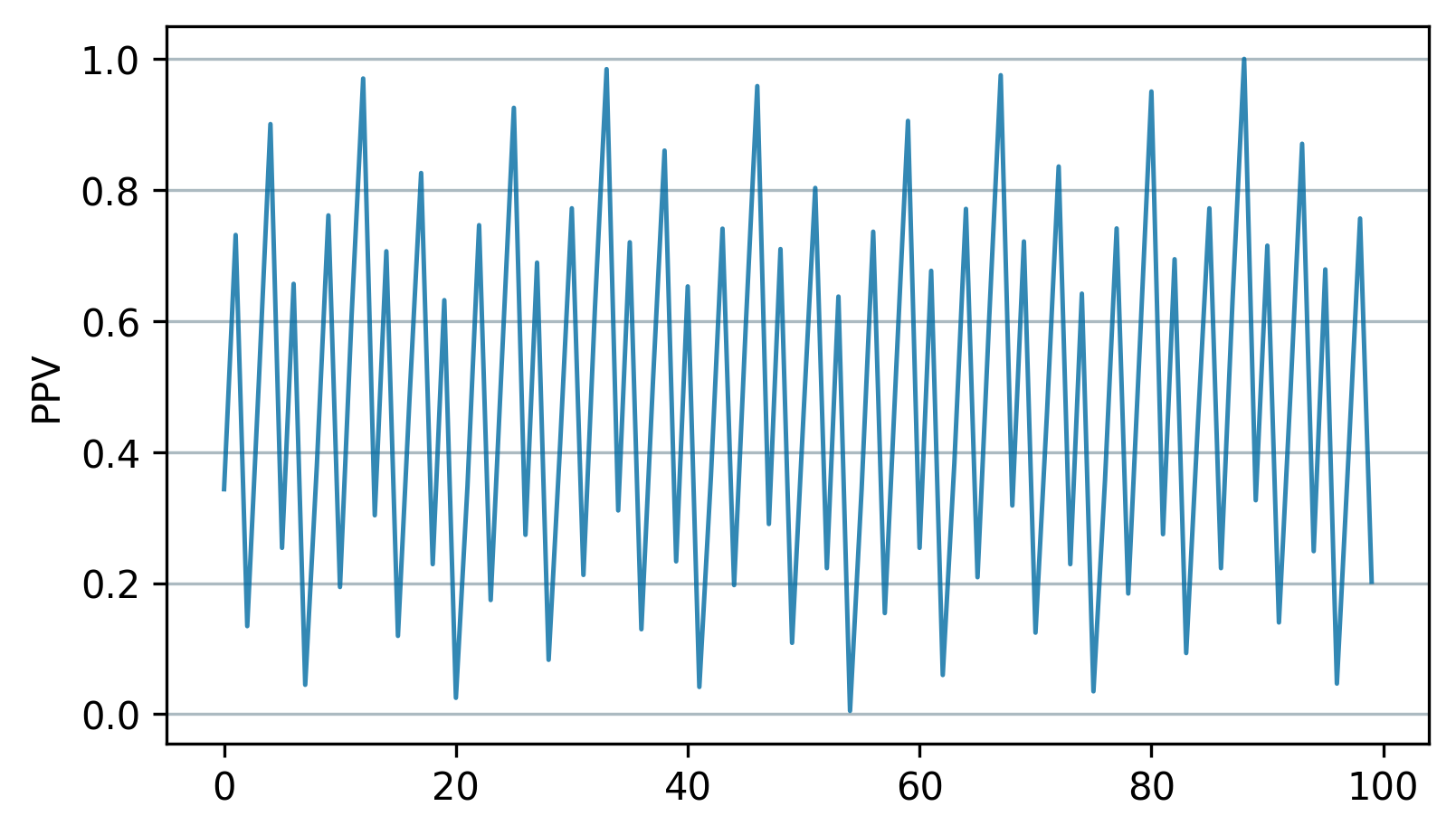}
    \caption{Transformed time series with the original Minirocket algorithm}
    \label{fig:transformed_fungi_original}
  \end{subfigure}
  \hfill
  \begin{subfigure}[b]{0.47\linewidth}
    \includegraphics[width=\textwidth]{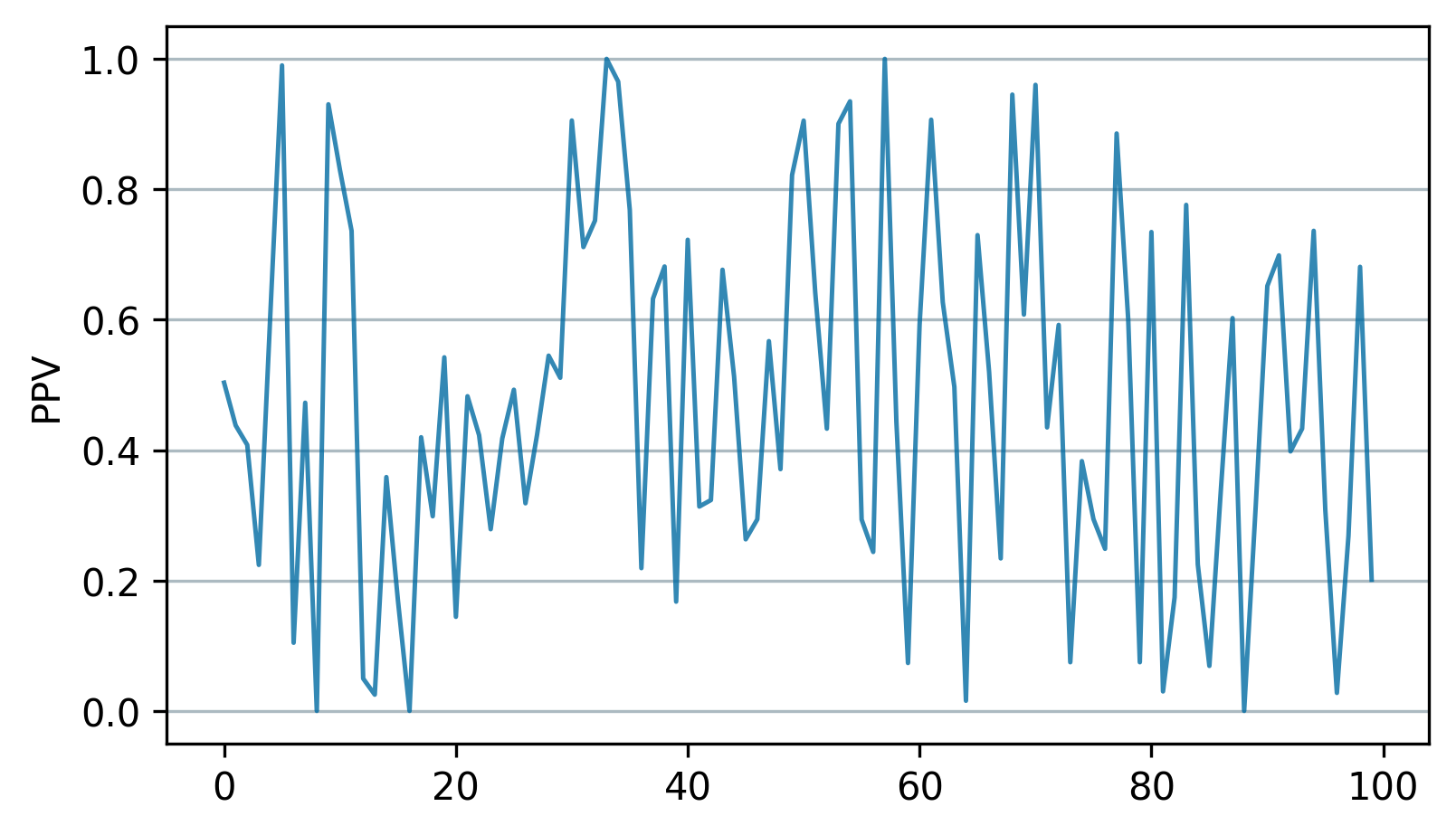}
    \caption{Transformed time series with the modified feature extractor}
    \label{fig:transformed_fungi_modified}
  \end{subfigure}
  \caption{Transformed sample time-series from Fungi dataset (figure \ref{fig:fungi-sample}) with the original feature extractor from Minirocket and the modified feature extractor}
  \label{fig:transformed_fungi}
\end{figure*}

Figure \ref{fig:fungi-sample} shows a sample time series from the UCR archive, while figure \ref{fig:transformed_fungi_original} illustrates its transformation through the original feature extractor, highlighting evident autocorrelations. To analyze these autocorrelation properties, we employ the Ljung-Box test, whose null hypothesis is that no autocorrelations exist at a specified lag. As expected from observing figure \ref{fig:transformed_fungi} (left), the test rejects the null hypothesis for every lag at 95\% confidence level, therefore we assume the presence of autocorrelations. To find out whether these periodic properties originate in the time series data, we introduce noisy data in the feature extractor, repeat the Ljung-Box test and still observe autocorrelations at every lag. This observation leads us to conclude that the original feature extractor introduces artificial autocorrelations, which are not inherent to the input time series. After modifying the algorithm as explained in Section \ref{subsec:feature_extraction} and introducing the same noise, we rerun the Ljung-Box test. In contrast to the previous findings, the test does not reject the null hypothesis at any lag. Thus, we assume the absence of autocorrelations, indicating that the modified feature extractor works as intended, that is producing noisy data from noisy inputs. For a sample transformation of a time series with the new updated feature extractor see Figure \ref{fig:transformed_fungi} (right)


\subsection{Search for Optimal Hyperparameters}
\label{subsec:extractor-configuration}
We explore several configurations of the algorithm and the effect on clustering accuracy. We focus on the main hyperparameters, specifically the kernel length and the number of kernels. For comparison, we set the number of kernels to range from 100 to 20,000 and kernel lengths to range from 7 to 13 and test various combinations across these entire ranges. We evaluated 20 combinations of hyperparameters on the development set of datasets. Among these, the combination of 500 kernels of length 9 yielded the highest number of wins (6), the best mean rank (8.12), and the second-highest mean accuracy (0.31), as shown in Table \ref{hyperparameter}. Based on these results, we selected the configuration of 500 kernels of length 9. Not only did this configuration obtains top positions in two of three categories, but its use of 500 kernels also made it faster than competitors utilizing 1000 or more kernels.

\begin{table}[ht]
\centering
\caption{Results for different hyperparameters configurations across the development set. The first term represents the number of kernels and the second the kernels's length. For instance, '500-9' stands for a configuration with 500 kernels of length 9}
\begin{tabular}{l@{\hspace{0.3cm}}c@{\hspace{0.3cm}}c@{\hspace{0.3cm}}c}
\toprule
\textbf{Algorithm} & \textbf{Mean rank} & \textbf{Mean accuracy} & \textbf{Winning count}  \\
\midrule\midrule
500-9           &       \textbf{8.12} &     0.310 &               \textbf{6} \\[3pt]
1000-11         &       8.17 &     0.299 &               2 \\[2pt]
10000-9         &       8.61 &     0.306 &               2 \\[2pt]
500-11          &       8.92 &     0.296 &               0 \\[2pt]
1000-9          &       9.08 &     \textbf{0.314} &               1 \\[2pt]
10000-7         &       9.14 &     0.302 &               1 \\[2pt]
5000-11         &       9.18 &     0.303 &               0 \\[2pt]
5000-9          &       9.24 &     0.305 &               3 \\[2pt]
10000-11        &       9.62 &     0.300 &               0 \\[2pt]
1000-7          &       9.83 &     0.295 &               1 \\[2pt]
5000-7          &       9.86 &     0.297 &               0 \\[2pt]
1000-13         &      10.22 &     0.286 &               1 \\[2pt]
500-7           &      10.44 &     0.284 &               1 \\[2pt]
500-13          &      10.68 &     0.280 &               1 \\[2pt]
5000-13         &      11.22 &     0.284 &               0 \\[2pt]
10000-13        &      11.29 &     0.286 &               1 \\[2pt]
100-9           &      12.07 &     0.256 &               3 \\[2pt]
100-11          &      14.08 &     0.242 &               1 \\[2pt]
100-13          &      14.61 &     0.226 &               3 \\[2pt]
100-7           &      15.60 &     0.214 &               1 \\[2pt]
\bottomrule
\end{tabular}
\label{hyperparameter}
\end{table}

\subsection{Dataset-level assessment}
We evaluate R-clustering on the validation dataset from the UCR archive using the same procedures as  in \cite{javed2020benchmark}. As the problem of finding the optimal partition of n data points into k clusters is an NP-hard problem with a non-convex loss function, we run the algorithm multiple times with different randomly initialized centroids to avoid local minima and enhance performance. Specifically, in line with the procedure adopted in the benchmark study, the algorithm is executed ten times, and the highest ARI score is chosen from all runs. The K-means algorithm requires the number of clusters to be specified in advance. While different methods, such as the elbow method \cite{thorndike1953belongs}, are available to estimate this number, evaluating these methods is not part of this benchmark. Given that we are using a labelled dataset in this experiment, the actual number of clusters is provided as input to the R-clustering algorithm.



\subsection{Results of the R-clustering algorithm}
\label{rclustering_results}
We compare the performance of R-clustering to the other algorithms of the benchmarks over the validation datasets under several perspectives using the ARI metric. We count the number of wins considering that ties do not sum and calculate the mean score and rank. R-clustering obtains:
\begin{itemize}
\item the highest number of wins (33) followed by Agglomerative (13) (see table \ref{winning_counts})

\item the best mean rank (3.47) followed by k-means-DTW (4.47) (see table \ref{mean_rank})

\item the best mean ARI score (0.324) followed by Agglomerative (0.276) (see tabgle \ref{aris}).
\end{itemize}

\begin{table}[h]
\centering
\caption{number of wins for each algorithm over the validation datasets in terms of best ARI}
\begin{tabular}{l@{\hspace{2cm}}c}
\toprule
\textbf{Algorithm} & \textbf{Winning count} \\
\midrule\midrule
R-Clustering             &  33 \\[3pt]
Agglomerative (Euclidean) &  13 \\[3pt]
K-shape                   &  10 \\[3pt]
Density Peaks (DTW)       &   5 \\[3pt]
K-means (DTW)             &   4 \\[3pt]
K-means (Euclidean)       &   3 \\[3pt]
C-means (Euclidean)       &   2 \\[3pt]
K-medoids (Euclidean)     &   1 \\[3pt]
Density Peaks (Euclidean) &   1 \\[3pt]
\bottomrule
\end{tabular}
\label{winning_counts}
\end{table}

\begin{table}[h]
\centering
\caption{Mean rank for R-Clustering and the rest of the algorithms considered in the benchmark}
\begin{tabular}{l@{\hspace{2cm}}c}
\toprule
\textbf{Algorithm} & \textbf{Mean rank} \\
\midrule\midrule
R-Clustering              &  3.47 \\[3pt]
Agglomerative (Euclidean) &  4.47 \\[3pt]
K-means (Euclidean)       &  4.49 \\[3pt]
K-means (DTW)             &  4.60 \\[3pt]
K-shape                   &  4.84 \\[3pt]
C-means (Euclidean)       &  5.20 \\[3pt]
K-medoids (Euclidean)     &  5.51 \\[3pt]
Density Peaks (Euclidean) &  5.89 \\[3pt]
Density Peaks (DTW)       &  6.53 \\[3pt]
\bottomrule
\end{tabular}
\label{mean_rank}
\end{table}

\begin{table}[h]
\centering
\caption{Average ARI for each algorithm over the validation datasets}
\begin{tabular}{l@{\hspace{2cm}}c}
\toprule
\textbf{Algorithm} & \textbf{Average ARI} \\
\midrule\midrule
R-Clustering              &  0.324 \\[3pt]
Agglomerative (Euclidean) &  0.276 \\[3pt]
K-means (Euclidean)       &  0.255 \\[3pt]
K-means (DTW)             &  0.248 \\[3pt]
C-means (Euclidean)       &  0.238 \\[3pt]
K-medoids (Euclidean)     &  0.232 \\[3pt]
K-shape                   &  0.206 \\[3pt]
Density Peaks (Euclidean) &  0.201 \\[3pt]
Density Peaks (DTW)       &  0.165 \\[3pt]
\bottomrule
\end{tabular}
\label{aris}
\end{table}


In addition, we perform a statistical comparison of R-clustering with the rest of the algorithms, in line with the recommendations provided by \cite{dau2019ucr} and \cite{demvsar2006statistical}. These recommendations suggest comparing the rank of the classifiers on each dataset. Initially, we perform the Friedman test \cite{friedman1937use} at a 95\% confidence level, which rejects the null hypothesis of no significant difference among all the algorithms. According to \cite{benavoli2016should}, upon rejecting the null hypothesis, it becomes neccesary to identify the significant differences among the algorithms. To accomplish this, we conduct a pairwise comparison of R-clustering versus the other classifiers using the Wilcoxon signed-rank test at a 95\% confidence level. This test includes the Holm correction for the confidence level, which adjusts for family-wise error (the chance of observing at least one false positive in multiple comparisons). Table \ref{holm_correction} presents the p-values from the Wilcoxon signed-rank test comparing R-Clustering with each of the other algorithms, together with the adjusted alpha values. The statistical rank analysis can be summarized as follows: R-clustering emerges as the best-performing algorithm with an average rank of 3.47 as presented in Table \ref{mean_rank}. The pairwise comparisons using R-Clustering as control classifier indicate that R-clustering presents significant differences in terms of mean rank with all other algorithms.

\begin{table}[h]
\centering
\caption{Results of the Wilcoxon signed-rank test between R-Clustering and the rest of the algorithms in consideration. The left column indicates the algorithm R-Clustering is compared to, the second column indicates the p-value of the Wilcoxon signed-rank test and the third column the alpha value with the Holm correction at a 95\% confidence level}
\begin{tabular}{l@{\hspace{1cm}}c@{\hspace{1cm}}c}
\toprule
\textbf{Algorithm} & \parbox[c][2.5em]{1cm}{\centering\textbf{p-value}} & \parbox[c][2.5em]{2cm}{\centering\textbf{alpha w/ Holm correction}} \\
\midrule\midrule
Density Peaks (DTW)       &  0.000001 &    0.006250 \\ [3pt]
Density Peaks (Euclidean) &  0.000010 &    0.007143 \\[3pt]
K-shape                   &  0.000012 &    0.008333 \\[3pt]
K-medoids (Euclidean)     &  0.000048 &    0.010000 \\[3pt]
C-means (Euclidean)       &  0.000633 &    0.012500 \\[3pt]
K-means (Euclidean)       &  0.001235 &    0.016667 \\[3pt]
K-means (DTW)             &  0.004060 &    0.025000 \\[3pt]
Agglomerative (Euclidean) &  0.045667 &    0.050000 \\[3pt]
\bottomrule
\end{tabular}
\label{holm_correction}
\end{table}

To wrap up this subsection, we incorporate the performance results of the R-Clustering algorithm without the PCA stage (See Table \ref{R_without}). This step is taken to validate and understand the contribution made by the PCA stage to the overall performance of the algorithm. In addition to being faster, R-Clustering outperforms R-Clustering without the PCA stage. However, it's worth noting that R-Clustering without PCA still achieves significant results and secures the second position across all three measured magnitudes.

\begin{table}[h]
\centering
\caption{Results for R-Clustering, R-Clustering without PCA and the rest of the algorithms considered in the benchmark}
\begin{tabular}{l@{\hspace{0.1cm}}c@{\hspace{0.1cm}}c@{\hspace{0.1cm}}c}
\toprule
\textbf{Algorithm} & \textbf{Mean rank} & \textbf{Mean accuracy} & \textbf{Winning count}  \\
\midrule\midrule
R-Clustering              &       4.00 &     0.324 &              21 \\[2pt]
R-Clustering W/O PCA      &       4.07 &     0.294 &              17 \\[2pt]
Agglomerative (Euclidean) &       5.09 &     0.276 &               8 \\[2pt]
K-means (Euclidean)       &       5.13 &     0.255 &               3 \\[2pt]
K-means (DTW)             &       5.26 &     0.248 &               3 \\[2pt]
K-shape                   &       5.43 &     0.206 &               9 \\[2pt]
C-means (Euclidean)       &       5.88 &     0.238 &               2 \\[2pt]
K-medoids (Euclidean)     &       6.20 &     0.232 &               1 \\[2pt]
Density Peaks (Euclidean) &       6.66 &     0.201 &               1 \\[2pt]
Density Peaks (DTW)       &       7.30 &     0.165 &               5 \\[2pt]
\bottomrule
\end{tabular}
\label{R_without}
\end{table}

\subsection{Computation Time and Scalability}


In this subsection, we compare the computation time and scalability of the R-Clustering algorithm with the Agglomerative algorithm and K-Shape, which are the second and third best performers based on winning counts. These two algorithms represent diverse approaches to time clustering, with Agglomerative employing a hierarchical strategy, and K-Shape utilizing a shape-based approach. The total computational time across all 112 datasets is 7 minutes for R-clustering (43 minutes for R-Clustering without PCA), 4 hours 32 minutes for K-shape, and 8 minutes for the Agglomerative algorithm. Nevertheless, it is important to highlight that the time complexity of the Agglomerative algorithm is \begin{math} \mathcal{O}(n^2) \end{math}  \cite{de2004open}. This characteristic might pose computational challenges for larger datasets and may limit proper scalability, as the subsequent experiment will show.

\begin{figure*}[t]
  \begin{subfigure}[b]{0.49\textwidth}
    \includegraphics[width=\textwidth]{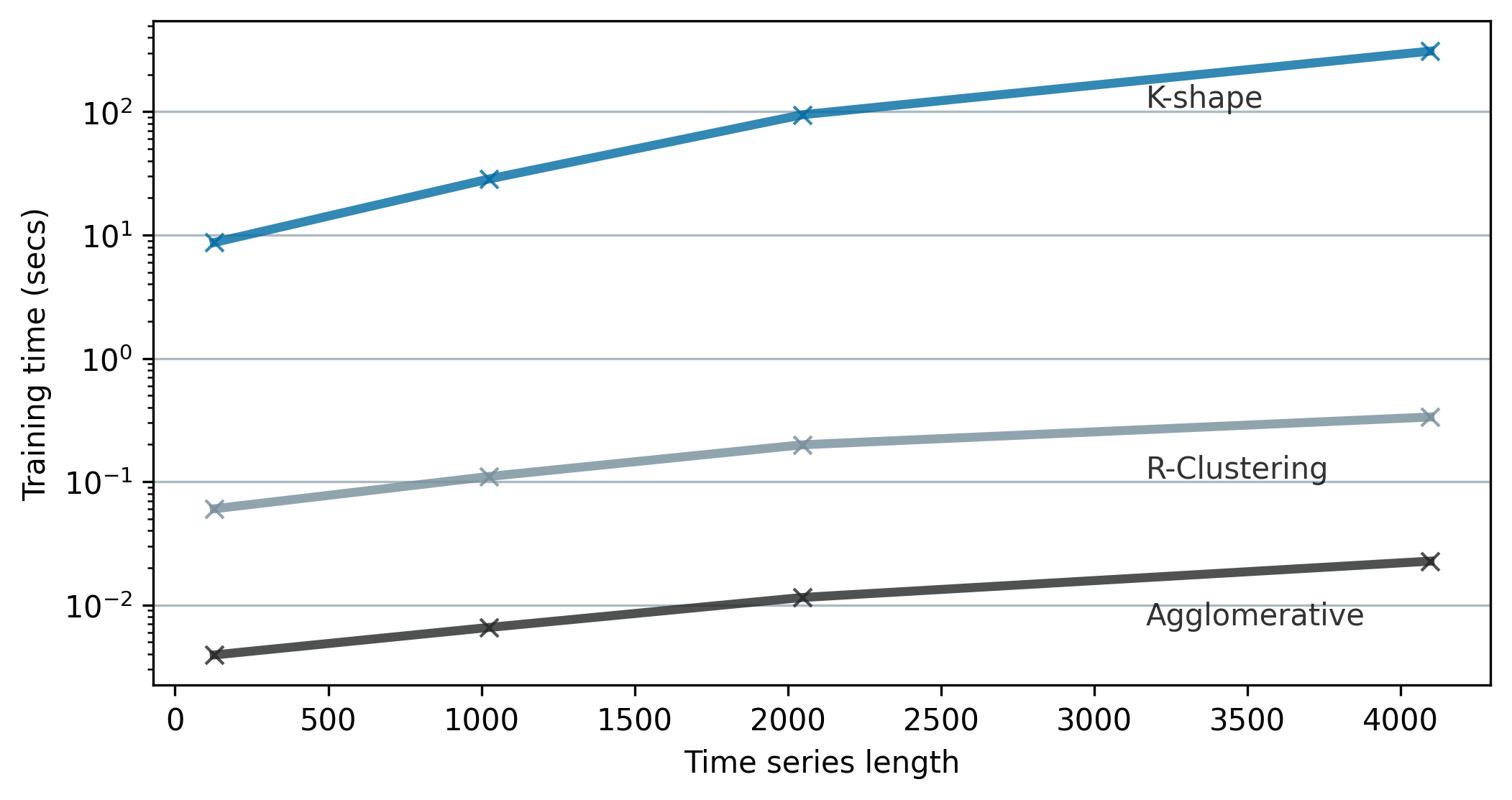}
    \caption{Performance on the DucksAndGeese dataset. The dataset size is fixed at 100 time series, with time series lengths varying. The graph depicts how changes in time series length impact the efficiency of the three algorithms}
    \label{fig:scalability_l}
  \end{subfigure}
  \hfill
  \begin{subfigure}[b]{0.49\textwidth}
    \includegraphics[width=\textwidth]{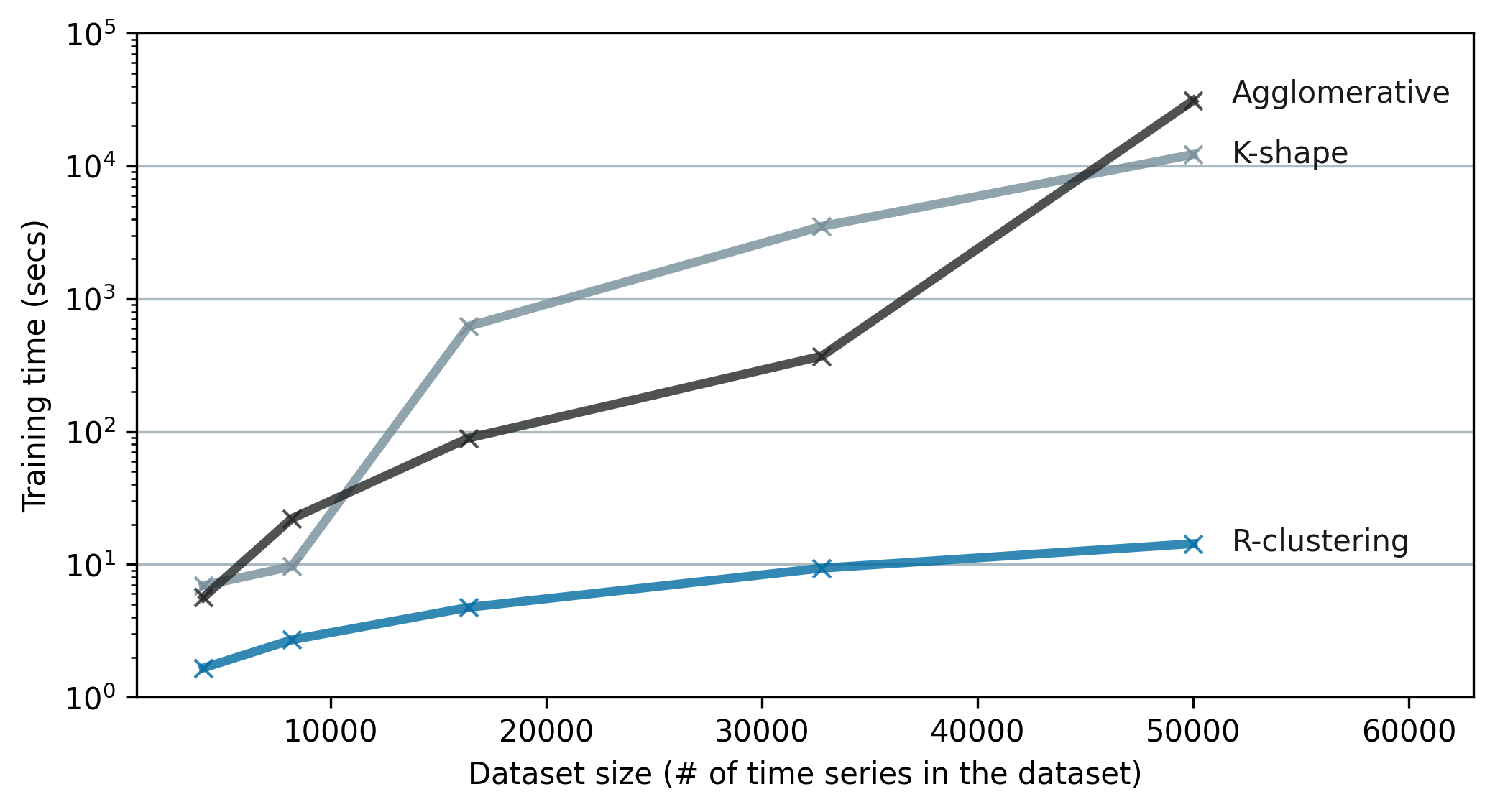}
    \caption{Comparative performance of the algorithms on the InsectSound dataset. The time series length is fixed at 600 points, while the size of the dataset varies. The graph illustrates the impact of changes in dataset size on the efficiency of the three algorithms}
    \label{fig:scalability_s}
  \end{subfigure}
  \caption{Scalability comparison of R-Clustering and the second and third-best performing algorithms with respect to time series length (left) and the number of time series in the dataset (right).}
  \label{fig:scalability}
\end{figure*}

In the scalability study, we use two recent datasets not included in the benchmark: DucksAndGeese and InsectSound. DucksAndGeese dataset consists of 100 time series across 5 classes, making it the longest dataset in the archive with a length of 236,784 points. The InsectSound dataset comprises 50,000 time series, each with a length of 600 points and spread across 5 classes. The results are depicted in figure \ref{fig:scalability}, which demonstrates the scalability of R-clustering in terms of both time series length and size. R-clustering scales linearly with respect to two parameters: the length of the time series and the number (or size) of time series in the dataset.  For smaller dataset sizes, Agglomerative algorithm performs the fastest, even for long time series. However, R-clustering outperforms the other algorithms when dealing with moderate to large datasets.

Despite the fact that the training stage of the Agglomerative algorithm is faster for certain data sizes, it exhibits drawbacks in some applications. R-clustering, like other algorithms based on K-means, can classify new data points easily using the centroids calculated during the training process. This is accomplished by assigning the new instance to the class represented by the nearest centroid. In contrast, the Agglomerative algorithm does not generate any parameter that can be applied to new instances. Consequently, when using Agglomerative to classify new data, the entire training process must be repeated, incorporating both the training data and the new observation.


\subsection{Statistical analysis of the benchmark}
To strengthen the results of the cited benchmark, in accordance with the recommendations provided by \cite{garcia2008extension}, we repeat the pairwise comparisons among each of the algorithms in the benchmark, not only with the newly presented method as a control classifier. The results are displayed in table \ref{p_values}, which indicates which pairs of algorithms exhibit a significant difference in performance regarding mean rank at a 95\% confidence level. It is important to note that the threshold for alpha value is not fixed at 0.05, but it is adjusted according to the Holm correction to manage the family-wise error \cite{demvsar2006statistical}. In this comparison, we notice that R-Clustering doesn't exhibit significant differences with certain algorithms as it did in the earlier comparison where it was the control classifier. The reason for this difference is the increased number of pairwise tests being conducted, which, in turn, diminishes the overall statistical power of the experiment.

The authors of the UCR dataset recommend showcasing these types of comparisons through a critical differences diagram, grouping together the algorithms which exhibit no significant difference. However, in our study, the diagram is not particularly insightful due to the large number of resulting groups. Therefore, instead of the diagram, we present the results in Table \ref{p_values}, which illustrates the significant differences between each pair of algorithms.

\section{Conclusions and future work}
\label{sec:discussion}
We have presented R-clustering, a clustering algorithm that incorporates static random convolutional kernels in conjuction with Principal Component Analysis (PCA). This algorithm transforms time series into a new data representation by first utilizing kernels to extract relevant features and then applying PCA for dimensionality reduction. The resultant embedding serves as an input for a K-means algorithm with Euclidean distance. We evaluated this new algorithm adhering to the procedures of a recent clustering benchmark study that utilizes the UCR archive - the largest public archive of labeled time-series datasets - and state-of-the-art clustering algorithms. Notably, R-clustering obtains the first place among all the evaluated algorithms across all datasets, using the same evaluation methods deployed in the reference study. Furthermore, we demonstrate the scalability of R-clustering concerning both time series length and dataset size, becoming the fastest algorithm for large datasets. Finally, in alignment with recent recommendations from the machine learning academic community, we strengthen the cited benchmark results with a pairwise statistical comparison of the included algorithms. This statistical analysis, coupled with the fact that the code used for generating the results in this paper is publicly available, should facilitate testing future time series clustering algorithms.

The effectiveness of random kernels in improving the clustering accuracy of time series has been demonstrated through the experimental results. This finding opens up several future research directions, such as adapting R-clustering for multivariate series, extending its use to other types of data like image clustering, and investigating the relationship between the number of kernels and performance. We anticipate that such a study could potentially suggest an optimal number of kernels depending on the time series length.

The excellent performance of the convolution operation with static random kernels has been demonstrated through experimental results. We, therefore, also encourage the academic community to engage in a more detailed analysis of the theoretical aspects of random kernel transformations. Progress in this direction could enhance our understanding of the process.

In conclusion, the R-clustering algorithm has shown promising results in clustering time series data. Its incorporation of static random convolutional kernels and PCA, along with its scalability and superior performance, make it a valuable tool for various applications. Future research and analysis in this field will contribute to the advancement of time series clustering algorithms and our understanding of random kernel transformations.



\bibliographystyle{IEEEtran}
\bibliography{bibliography}



\newpage

\begin{table*}[h]
\centering

\begin{tabular}{lllc}
\toprule
              Algorithm 1 &               Algorithm 2 &      p value &  Significant difference \\
\midrule
Density Peaks (Euclidean) &                       R-clustering & 2.939982e-08 &                    True \\
      K-means (Euclidean) &     K-medoids (Euclidean) & 7.516924e-08 &                    True \\
      Density Peaks (DTW) &                       R-clustering & 3.651556e-07 &                    True \\
Agglomerative (Euclidean) &     K-medoids (Euclidean) & 1.183446e-06 &                    True \\
    K-medoids (Euclidean) &                       R-clustering & 6.988398e-06 &                    True \\
      Density Peaks (DTW) &             K-means (DTW) & 9.209253e-06 &                    True \\
Agglomerative (Euclidean) &       Density Peaks (DTW) & 1.863177e-05 &                    True \\
Density Peaks (Euclidean) &       K-means (Euclidean) & 4.440957e-05 &                    True \\
      Density Peaks (DTW) &       K-means (Euclidean) & 7.363479e-05 &                    True \\
      C-means (Euclidean) &                       R-clustering & 9.099955e-05 &                    True \\
Agglomerative (Euclidean) & Density Peaks (Euclidean) & 1.541735e-04 &                    True \\
      K-means (Euclidean) &                       R-clustering & 1.910857e-04 &                    True \\
Density Peaks (Euclidean) &             K-means (DTW) & 2.032777e-04 &                    True \\
      Density Peaks (DTW) &     K-medoids (Euclidean) & 5.362103e-04 &                    True \\
Agglomerative (Euclidean) &       K-means (Euclidean) & 6.785202e-04 &                    True \\
            K-means (DTW) &     K-medoids (Euclidean) & 6.857564e-04 &                    True \\
            K-means (DTW) &                       R-clustering & 1.207988e-03 &                    True \\
Agglomerative (Euclidean) &       C-means (Euclidean) & 1.264458e-03 &                    True \\
Density Peaks (Euclidean) &                   K-shape & 1.682662e-03 &                    True \\
      C-means (Euclidean) &       K-means (Euclidean) & 2.149903e-03 &                    True \\
      C-means (Euclidean) &             K-means (DTW) & 2.312641e-03 &                    True \\
                  K-shape &                       R-clustering & 4.440235e-03 &                   False \\
      Density Peaks (DTW) &                   K-shape & 4.906171e-03 &                   False \\
Agglomerative (Euclidean) &                       R-clustering & 1.117005e-02 &                   False \\
Density Peaks (Euclidean) &     K-medoids (Euclidean) & 2.044118e-02 &                   False \\
      C-means (Euclidean) &       Density Peaks (DTW) & 5.935112e-02 &                   False \\
      Density Peaks (DTW) & Density Peaks (Euclidean) & 1.089868e-01 &                   False \\
    K-medoids (Euclidean) &                   K-shape & 1.367893e-01 &                   False \\
      C-means (Euclidean) &                   K-shape & 1.850786e-01 &                   False \\
      C-means (Euclidean) & Density Peaks (Euclidean) & 2.050779e-01 &                   False \\
            K-means (DTW) &       K-means (Euclidean) & 2.188934e-01 &                   False \\
            K-means (DTW) &                   K-shape & 2.249359e-01 &                   False \\
Agglomerative (Euclidean) &                   K-shape & 3.193606e-01 &                   False \\
Agglomerative (Euclidean) &             K-means (DTW) & 7.882831e-01 &                   False \\
      C-means (Euclidean) &     K-medoids (Euclidean) & 9.768389e-01 &                   False \\
      K-means (Euclidean) &                   K-shape & 9.918929e-01 &                   False \\
\bottomrule
\end{tabular}

\caption{Pairwise comparisons of the algorithms according to the Wilcoxon signed-rank test with a 95\% confidence level and Holm correction applied to the alpha values. }
\label{p_values}
\end{table*}

\vfill

\end{document}